\documentclass{llncs}
\usepackage{booktabs,caption,fixltx2e}
\usepackage[flushleft]{threeparttable}
\usepackage{graphicx}
\usepackage{makeidx}  
\usepackage{amsmath}
\usepackage{multirow}

\usepackage{amsfonts}%
\usepackage{amssymb}%
\usepackage{graphics}
\usepackage{subfig}
\usepackage[labelfont=bf]{caption}
\usepackage[font={small}]{caption}
\usepackage{transparent}
\usepackage{color}
\usepackage[ruled,vlined]{algorithm2e}
\usepackage{tabularx}
\usepackage{rotating}
\usepackage{makecell}
\usepackage{enumitem}
\usepackage{mathtools}

\begin{document}

\title{Cross-Modal Manifold Learning (C$\text{M}^{2}$L) for Medical Image Retrieval}  

%
\author{Sailesh Conjeti\inst{1*}\and Anees Kazi \inst{1}\thanks{S. Conjeti and A. Kazi contributed equally towards the work.} \and Nassir Navab\inst {1,4} \and Amin Katouzian\inst{3}}
\authorrunning{S. Mesbah et al.} 
\institute{
Computer Aided Medical Procedures, Technische Universit\"{a}t M\"{u}nchen, Germany.\\
\and
IBM Almaden Research Center, San Jose, CA, USA.\\ 
\and
Computer Aided Medical Procedures, Johns Hopkins University, USA.}

\maketitle              
\begin{abstract}
This paper presents a new scalable algorithm for cross-modal similarity preserving retrieval in a learnt manifold space. Unlike existing approaches that compromise between preserving global and local geometries, the proposed technique respects both simultaneously during manifold alignment. The global topologies are maintained by recovering underlying mapping functions in the joint manifold space by deploying partially corresponding instances. The inter-, and intra-modality affinity matrices are then computed to reinforce original data skeleton using perturbed minimum spanning tree (pMST), and maximizing the affinity among similar cross-modal instances, respectively. The performance of proposed algorithm is evaluated upon two multimodal image datasets (coronary atherosclerosis histology and brain MRI) for two applications: classification, and regression. Our exhaustive validations and results demonstrate the superiority of our technique over comparative methods and its feasibility for improving computer-assisted diagnosis systems, where disease-specific complementary information shall be aggregated and interpreted across modalities to form the final decision.

\end{abstract}

\section{Introduction}

Multi-modal imaging is increasingly performed to obtain disease-specific complementary data throughout diagnosis-treatment-follow-up procedures. A reliable cross-modal image retrieval system is desirable as it carries immense potential in aiding decision-making by enabling access to all information across modalities that share semantic similarity. In contrast to single-modal image retrieval that has been an active research topic~\cite{zhang2015}, the cross modal image retrieval has not yet been fully investigated for medical applications except few works~\cite{kumar2008},~\cite{Cao2011},~\cite{kalpathy2015} that are also mainly adopted for health care management systems using text+image datasets. In cross-modal retrieval task, the ultimate goal is to bridge the gap between feature spaces by mining their mutual correlations and unveiling similarities within a latent common space. To this end, several methods have been developed but the Canonical Correlation Analysis (CCA)~\cite{hotelling1936} and its variants (ex.~\cite{Rasiwasia2010}) have been widely used for learning such a space by maximizing the correlation between the two feature spaces. Alternatively, learning coupled feature spaces (LCFS)~\cite{wang2013LCFS} and Procrustes alignment~\cite{wang2013} algorithms have been proposed, where the former focuses on selecting discriminative features while learning the subspace and the latter removes translational, rotational, and scaling components from one space so that the optimal alignment can be achieved. In general, majority of existing methods only preserve local geometries amongst features and ignore global geometries. In other words, they only ensure that similar instances in the original space become neighbors in the latent space but do not prevent dissimilar instances from being neighbors. Authors in~\cite{wang2013} addressed this problem by projecting instances into latent space through recovered mapping functions upon partially corresponding instances and aligning the manifolds while preserving the global geometries. 

The data across imaging modalities are inherently heterogeneous due to differences in physics of acquisitions and protocols. Therefore, we need to preserve local structures that carry information about population variability and at the same time preserve the global manifold geometries for coping up with disease heterogeneity. This motivated us to develop the cross-modal manifold learning (C$\text{M}^{2}$L) image retrieval algorithm that respects both geometries in the latent space, which is accounted as the main contribution of this paper. This is achieved by: 1) incorporating pMST~\cite{zemel2004} into C$\text{M}^{2}$L such that the original data skeleton is preserved and partially corresponding instances drive the alignment, and 2) introducing novel  notion of proximity through inter- and intra-modality affinity matrices for maintaining local similarities within constructed graph neighborhood. To the best of our knowledge, this is the first work on cross-modal medical image retrieval and as proof of concept, we apply C$\text{M}^{2}$L on two datasets for two different applications. For \textit{classification}, we deployed coronary atherosclerosis histology images stained with Hematoxylin and Eosin (HnE) and Movat Pentachrome (MP). Each stain provides distinctive information about atherosclerotic tissues that are interpreted in search of vulnerable plaques~\cite{virmani2000}, which has high clinical significance. The need for cross-modal retrieval arises when either of two stains are available (\textit{query}) and the histopathologist desires access to the complementary stained image for differential assessment and tissue labeling that is otherwise very tedious and time consuming. Ideally, the retrieved regions of interest should share the same semantic similarity in terms of the disease and arterial morphology as that of the query. For \textit{regression}, we used BraTS dataset, which comprises of 3D scans of 188 High Grade Gliomas (HGG) and 25 Low Grade Gliomas (LGG) acquired with four MRI contrast schemes: T1, TIc, T2, and FLAIR (Refer to~\cite{menze2015} for more details). We pose the task of cross-modal retrieval for scenarios where any triad of the above four modalities are available (\textit{query}) and the neurologist seeks the other complementary modality for improved diagnosis and treatment planning. Ideally, the retrieved cross-modal volume/region should share similarity in tumor composition to that of the query triad and similarity in the tumor staging.

\section{Methodology}

 \begin{figure}[t]
	\centering
		\includegraphics[width=0.9\textwidth]{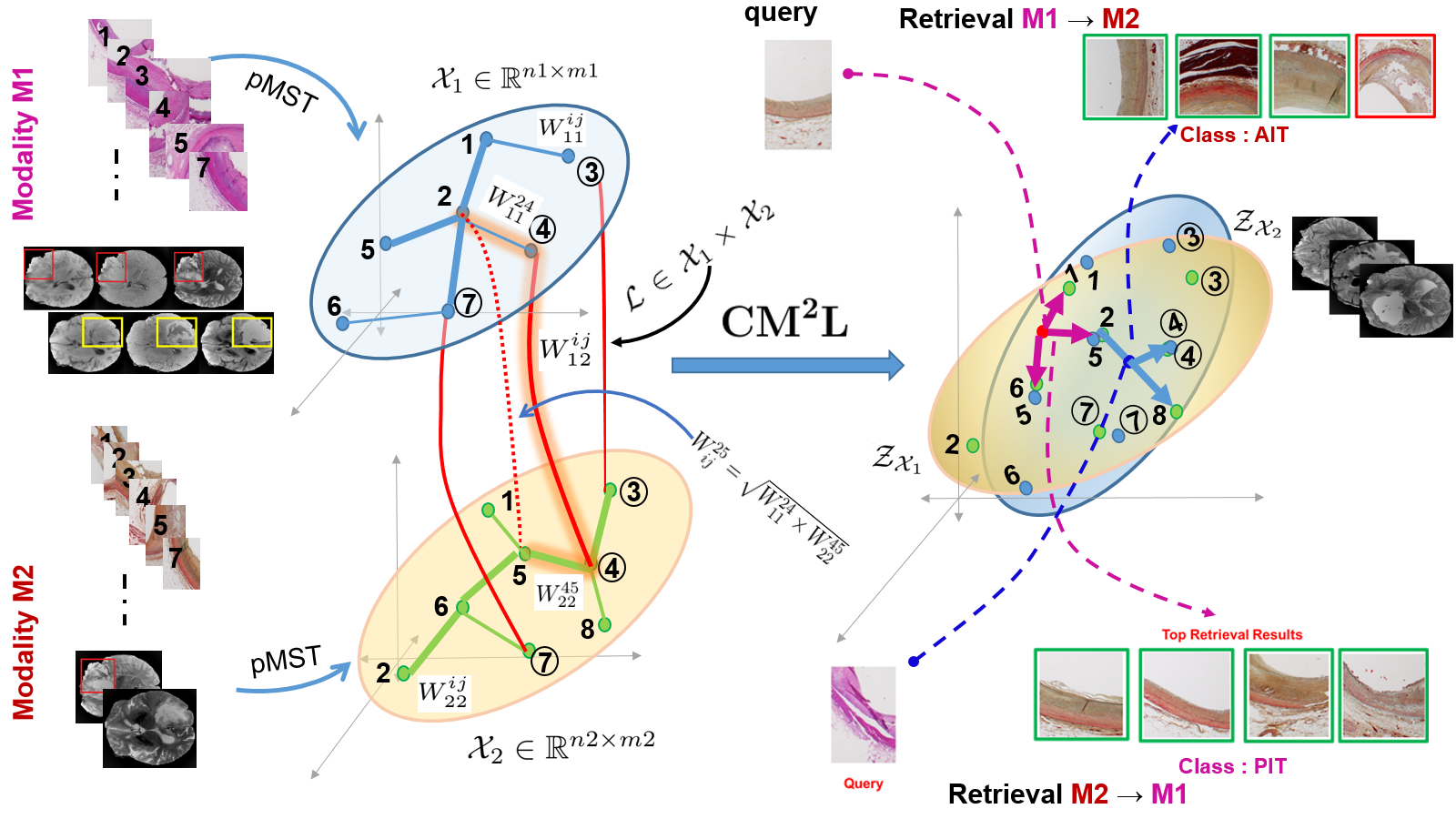}
	\caption{Retrieval schematic of C$\text{M}^{2}$L: Given modalities $\mathcal{X}_{1}$, $\mathcal{X}_{2}$, and limited corresponding instances ($\mathcal{L}$), we model the intra-modal proximity with pMST. We then leverage $\mathcal{L}$ through C$\text{M}^{2}$L and learn to map them to the joint feature space $\mathcal{Z}: \mathcal{Z}_{\mathcal{X}_{1}} \cup \mathcal{Z}_{\mathcal{X}_{2}}$.}
	\label{fig:overall}
\end{figure}

Given two sets of multi-modal images $\mathcal{X}_{1} = \left \{ \mathbf{x}_{1}^{i} \in \mathbb{R}^{m_{1}} \right \}_{i = 1} ^{n_{1}}$  and $  \mathcal{X}_{2} = \left \{ \mathbf{x}_{2}^{i} \in \mathbb{R}^{m_{2}} \right \}_{i = 1} ^{n_{2}}$, as a prerequisite, we collect $n_{\mathcal{L}}$ number of partially corresponding data (tuples) from both modalities constituting $\mathcal{L} \in \mathcal{X}_{1} \times \mathcal{X}_{2}$. We reconstitute $\mathcal{X}_{1}$ into two disjoint subsets: $\mathcal{X}_{1}^{c}$ and $\mathcal{X}_{1}^{wc}$ (with and without given tuples, respectively) and likewise partition $\mathcal{X}_{2}$. 
In presence of limited correspondences, just preserving neighborhood relationships amongst matching instances would result in overfitting, thus limiting generalization. C$\text{M}^{2}$L overcomes this by using $\mathcal{X}_{1}^{wc}$ and $\mathcal{X}_{2}^{wc}$ together with $\mathcal{L}$ such that the whole global geometries of two underlying manifolds are coupled and aligned in the joint feature space.

The C$\text{M}^{2}$L casts the retrieval problem into learning a latent metric $q$-dimensional space $\mathcal{Z} = \left ( \mathcal{Z}_{\mathcal{X}_{1}} \cup \mathcal{Z}_{\mathcal{X}_{2}} |  \mathcal{X}_{1} \rightarrow \mathcal{Z}_{\mathcal{X}_{1}};  \mathcal{X}_{2} \rightarrow \mathcal{Z}_{\mathcal{X}_{2}}  \right ) \in R^{q}$  wherein $\mathcal{X}_{1}$ and $\mathcal{X}_{2}$ become comparable ($q\leq \min(m_{1},m_{2})$). 

The overall geometry encompassing both intra- and inter-modal global geometries of the aligning manifolds can be model as a $(n_1+n_2) \times (n_1+n_2)$ joint distance matrix $\mathcal{D}$ representing the pairwise dissimilarity between any two instances in $\left \{ \mathcal{X}_{1},\mathcal{X}_2 \right \}$. The joint manifold geometry is defined as $\tau \left ( \mathcal{D} \right ) = -HSH/2$, where $S_{ij} = D^{2}_{ij}$, $H_{ij} = \mathcal{I}^{(n_1+n_2)\times(n_1+n_2)} - (1/(n_1+n_2))$, and $\mathcal{I}$ is an identity matrix (construction of $\mathcal{D}$ is discussed later)~\cite{tenenbaum2000}. $\mathcal{Z}$ is estimated through: 
\begin{equation} 
(\mathcal{Z}_{\mathcal{X}_{1}}^{*},\mathcal{Z}_{\mathcal{X}_{2}}^{*})  =  \underset{\mathcal{Z}_{\mathcal{X}_{1}},\mathcal{Z}_{\mathcal{X}_{2}}}{\arg\min}\left \| \tau \left ( \mathcal{D} \right ) -\left [ \mathcal{Z}_{\mathcal{X}_{1}},\mathcal{Z}_{\mathcal{X}_{2}} \right ]^{T} \left [ \mathcal{Z}_{\mathcal{X}_{1}},\mathcal{Z}_{\mathcal{X}_{2}} \right ]\right \|^{2}
\label{eq:Opt}
\end{equation}

Next, the intra- and inter-modal similarities shall be preserved in $\mathcal{Z}$. The former is discovered within modality neighborhoods and modeled as intra-modal pMSTs. The inter-modality neighborhoods are then inferred using $\mathcal{L}$, which act as `links' and aid in aligning the intra-modal pMSTs such that similar instances across modalities are mapped close to one another in the learnt manifold space ensuring exclusion of dissimilar points within local neighborhood (see Fig.~\ref{fig:overall}).

\noindent
\texttt{Step 1} \textit{\textbf{pMST}}: The minimum spanning tree (MST) of a data distribution effectively represents the underlying skeleton of a manifold, preserves local structures, and does not introduce \textit{gaps} between small random groupings of data points, which  guarantees connectedness of a graph. However, MST is too sparse and sensitive to noise and therefore a fully-connected graph is often constructed, resulting in erroneous connections traveling outside of an underlying manifold when maximizing inter-modality proximity. To resolve this issue, we employed pMST, which is an ensemble of MSTs applied on perturbed versions of the original data distribution (MST$(\mathcal{X}_{p})$), where $\mathcal{X}_{p} = \left \{ \mathbf{x}_{i}^{p} \mid \mathbf{x}_{i}^{p} \in \mathcal{N}(\mathbf{x}_{i}, \sigma_{i})\text{; }\mathbf{x}_{i} \in \mathcal{X}  \right \}$ and $\sigma_{i} = r_{p} \times d(\mathbf{x}_{i},\mathbf{x}_{i}^{k}); \text{ with } r_{p} \in [0,1]$ that is locally adaptive noise model~\cite{zemel2004}. The edge $e_{ij}^{p}$ between two points $\mathbf{x}_{i}^{p}$ and  $\mathbf{x}_{j}^{p}$ is 1 if they are connected in MST$(\mathcal{X}_{p})$ and 0 otherwise. The final edge weight between two $\mathbf{x}_{i}$ and $\mathbf{x}_{j}$ instances in the pMST neighborhood graph (say $\delta_{\mathcal{X}}$) is computed as $e_{ij} = \frac{1}{t_{p}} \sum _{p=1}^{t_{p}} e_{ij}^{p}$. Next, we define the intra-modal proximity graph, deploying these two constructed pMST graphs, $\delta_{\mathcal{X}_{1}}$ and $\delta_{\mathcal{X}_{2}}$ for $\mathcal{X}_{1}$ and $\mathcal{X}_{2}$ data points, respectively.

\noindent
\texttt{Step 2}\textit{\textbf{ Intra-modal Affinity}}: For defining proper affinity and evaluating similarity/dissimilarity between data points $\mathbf{x}_{i}$ and $\mathbf{x}_{j}$ in the manifold, we incorporate locally scaled $l_{2}$ norm into intra-modal distance metric $ \mathcal{D}_{ij} = \left \| \mathbf{x}_{i} - \mathbf{x}_{j} \right \|^{2}/ (2\sigma _{\mathbf{x}_{i}}{\sigma _{\mathbf{x}_{j}}})$, where $\sigma_{i}$ and $\sigma_{j}$ are local scaling factors measured by $\sigma_{i} = \left \| \mathbf{x}_{i} - \mathbf{x}_{K} \right \|^{2}$~\cite{zelnikmanor2004} such that $\mathbf{x}_{K}$ is the $K^{\text{th}}$ neighbor of $\mathbf{x}_{i}$. This allows for self-tuning of point-to-point distances in local neighborhoods around the points $\mathbf{x}_{i}$ and $\mathbf{x}_{j}$. In cross-modal retrieval, the heterogeneous gap between feature spaces warrants that we normalize the intra-modal distance matrices $\mathcal{D}_{11}$ and $\mathcal{D}_{22} $ to make them comparable. Finally, the intra-modal affinity is measured as $W_{11} = \text{exp}(-\mathcal{D}_{11})\text{. }\delta_{\mathcal{X}_{1}}$ for $\mathcal{X}_{1}$ and likewise for $\mathcal{X}_{2}$.

\noindent
\texttt{Step 3}\textit{\textbf{ Inter-modal Affinity}}: Given partial correspondences $\mathcal{L}$, the inter-modal affinity is derived by the inferred intra-modal affinities $W_{11}$, $W_{22}$. The corresponding instances across modalities are treated as `links' to leverage during alignment. For any pair of cross-modal points (say $\mathbf{x}_{1}^{i}$ and $\mathbf{x}_{2}^{j}$ ), the cross-modal affinity is computed as the maxima of affinities through all possible `links' between the modalities, \textit{i.e.} $W_{12}^{ij} = \underset{{k\in\left [ 1,n_{\mathcal{L}} \right ]}}{\text{max}} \sqrt{W_{11}^{ik}\times W_{22}^{kj}}$, where $W_{11}^{ik}$ is the intra-modal affinity between $\mathbf{x}_{1}^{i}$ and data point $\mathbf{x}_{1}^{k}$,  and $W_{22}^{kj}$ is the intra-modal affinity between $\mathbf{x}_{2}^{j}$ data point $\mathbf{x}_{2}^{k}$ where $\left ( \mathbf{x}_{1}^{k},\mathbf{x}_{2}^{k} \right ) \in \mathcal{L}$. The final composite distance matrix representing the joint geometry is computed as: $\mathcal{D} = 1- \begin{psmallmatrix*}W_{11} & W_{12} \\ 
W_{12}^{T}  & W_{22} \end{psmallmatrix*}$.

\noindent
\textbf{Learning Latent Space $\mathcal{Z}$ and Out of Sample Extension}: Solution to Eq.~\ref{eq:Opt} posed earlier is the eigen-decomposition of $\tau(\mathcal{D})$ as $\tau\left ( \mathcal{D} \right ) = U^{t}\text{diag}(\Lambda_{1},\cdots,\Lambda_{q} )U$ where $U \in \mathbb{R}^{(n_{1}+n_{2})\times q}$, and $\Lambda$ are eigenvalues, therefore, the latent subspace is estimated as $\mathcal{Z} = \text{diag}\left ( \Lambda_{1},\cdots,\Lambda_{q}\right )^{1/2}U$~\cite{tenenbaum2000}. 
For an unseen data point $\mathbf{x}_{t}$, we adapt the formulation from~\cite{strange2011}, which computes locally adaptive tangent spaces for out of sample extension (OSE). Through OSE, we seek the corresponding projected point $\mathbf{z}_{t} \in \mathcal{Z}$ and leverage the local neighborhood $N(\mathbf{x}_{t})$ defined in the high-dimensional space to search for a locally linear mapping function $M$ such that $\mathbf{z}_{t} = M\mathbf{x}_{t}$, where $M$ is decomposable into two piecewise matrices $A$ and $V$ ($M = AV$). We infer $V$ as the eigen-vectors corresponding to the top $q$ non-zero eigenvalues generated through Principal Components Analysis (PCA) on $N(\mathbf{x}_{t}) \cup\mathbf{x}_{t}$ and $A$ is the similarity transformation matrix (translation, scaling, and rotation) that is learnt through local Procrustes alignment.

\noindent
\textbf{Cross-modal Retrieval in Joint Manifold Space}: Through C$\text{M}^{2}$L, we make the projected spaces $\mathcal{Z}_{\mathcal{X}_{1}}$ and $\mathcal{Z}_{\mathcal{X}_{2}}$ metric-comparable. Therefore, without loss of generality, the task of cross-modal retrieval for a query (say, $\mathbf{x}_{q}$ of modality $M1$) will be casted as projecting it appropriately onto the joint space ($\mathbf{z}_{q} = \text{OSE}(\mathbf{x}_{q})$) and fetching the closest projected points from target modality ($\mathcal{Z}_{\mathcal{X}_{2}}$). 

\noindent
\textbf{Extension to Feature Level Alignment}: So far, the C$\text{M}^{2}$L was elaborated as it searches for and establishes non-linear mapping of original feature spaces and joint embedding space, which we refer to it as ``instance-level" version (C$\text{M}^{2}$L-I). It can be seamlessly generalized to the case of linear embedding by replacing $\mathcal{Z}_{\mathcal{X}_{1}}$ and $\mathcal{Z}_{\mathcal{X}_{2}}$ in Eq.~\ref{eq:Opt} with $\alpha^{t}\mathcal{X}_{1}$ and  $\beta^{t}\mathcal{X}_{2}$, respectively. The solution is given by the eigenvectors corresponding to the $q$ maximum non-zero eigenvalues of  $Z\tau \left ( \mathcal{D} \right )V^{T}\gamma  = \lambda VV^{T}\gamma$ where $V =  \begin{psmallmatrix*}
\mathcal{X}_{1} & 0^{n_{1}\times d_{2}}\\ 
0^{n_{2}\times d_{1}} & \mathcal{X}_{2} 
\end{psmallmatrix*}$ where $\gamma = [\alpha,\beta]$~\cite{wang2013}. This linear feature-level variant of C$\text{M}^{2}$L is thereafter refered to as C$\text{M}^{2}$L-F.

\section{Experiments and Results}
\textbf{Validation Scheme}: The performance of both C$\text{M}^{2}$L-I and C$\text{M}^{2}$L-F algorithms are evaluated against comparative methods as listed in Table~\ref{tab: comparativemethods}. 
We randomly split the data into two disjoint subsets with a 80:20 ratio corresponding to the training and test datasets and repeated the splitting 10 times. For robustness, we quantify the retrieval performance varying the degree of given correspondences for two settings of 20\% (sparse) and 80\% (dense) correspondences.

\begin{table}[t]\small
\centering
\caption{Comparative methods and their configurational settings}
\label{baselinesTable}
\resizebox{\textwidth}{!}{%
\begin{tabular}{c|c|c|cc}
\makecell{ \textbf{Methods} \textbf{with abbreviations}}& \textbf{Type}  & \textbf{Graph} & \textbf{Hyperparameters} \\
\hline\hline
 Cannonical Correlation Analysis (CCA)~\cite{hotelling1936}& F & $\times$ & Cross-modal Correlation $> 0.1$  \\
\hline
\makecell{Manifold alignment preserving global geometry (MA-F and MA-I)~\cite{wang2013} }&  F and I&  FC & \makecell{Eigen-value threshold $\epsilon > 10E-05$; $k$ for OSE = 20} \\
\hline
Learning coupled feature spaces (LCFS)~\cite{wang2013LCFS} & F & $\times$ & \makecell{Regularization parameters $\lambda_{1} = 10E-01$ ,$\lambda_{2}$ = 10E-03 \\ Number of iterations = 10}\\
\hline
Procrustus Alignment (PA) & $\times$ & $\times$ & -\\
\hline
Cross-Modal Manifold Learning (C$\text{M}^{2}$L)& F and I & pMST & \makecell{Number of perturbations $t_{p}= 20$ \\ Locally adaptive noise model $r_{p} = 0.5; k = 5$ \\ Eigen-value threshold $\epsilon > 10E-05$; $k$ for OSE = 20}\\
\hline
\hline
\end{tabular}
}
\label{tab: comparativemethods}
\end{table}

\noindent
\textbf{Staging of coronary atherosclerosis through cross-modal retrieval}:
We followed an acquisition protocol in~\cite{katouzian2011} and collected 253 HnE and MP pairs of cross-sections from 16 coronary arteries excised from 6 \textit{post-mortem} human hearts, resulting in 16467 regions of interest (ROIs) with variable sizes (between $ 640 \mu$m $\times$ $ 640 \mu$m and  $ 2560 \mu$m $\times$ $ 2560 \mu$m). The stains are performed on consecutive cross-sections ($< 5 \mu$m apart) and rigidly registered manually. Eleven Modified AHA~\cite{virmani2000} labels were used for annotations of underlying tissues in accordance with interpretations from an expert cardiovascular histopathologist. It must be noted that C$\text{M}^{2}$L does not use labels during training and these annotations are used purely for validations (This extends to BraTS dataset too). The ROIs were then fed into a pre-trained deeply learnt Convolutional Neural Network (CNN) trained for large-scale recognition tasks. We used outputs arising from the second fully connected layer (FC2) of VGG-F~\cite{chatfield2014} and AlexNet~\cite{krizhevsky2012} deep CNN networks as 4096-dimensional features for HnE and MP images, respectively. Two different networks were chosen to maintain the heterogeneous gap between the raw feature spaces, which would subsequently be bridged through C$\text{M}^{2}$L and comparative methods. Further, to make the feature spaces discriminative, we reduced dimensionality using supervised locally linear projections, preserving 90\% data variance~\cite{he2004}. 

\begin{figure}[t]
\centering
\includegraphics[width = \textwidth]{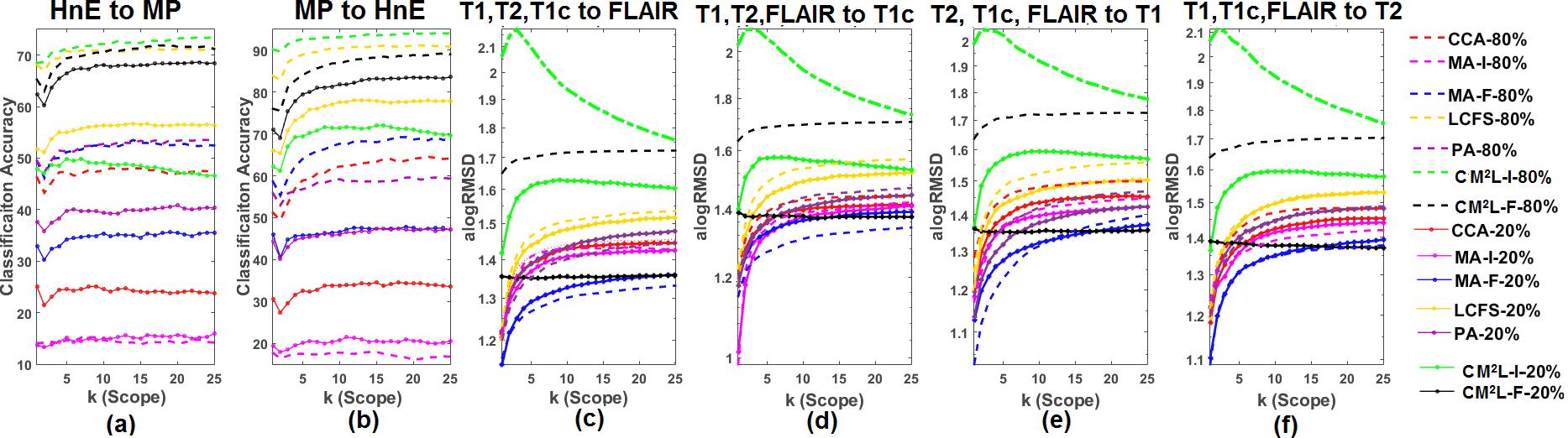}
\caption{Performance \textit{vs.} Scope ($k$ retrieved cross-modal neighbors) curves for the proposed and comparative methods for 20\% and 80\% degrees of correspondence.}
\label{fig:resultsKCurve}
\end{figure}

\noindent
\textit{Results}: The retrieval performance are measured using classification accuracy and for a particular query instance (belonging to either HnE or MP) the class is predicted as the maximum a posteriori class evaluated from the top $k$ nearest cross-modal neighbors. Fig.~\ref{fig:resultsKCurve} depicts the overall performance, varying $k$ through the accuracy-scope ($k$) curve for two settings of nearest neighbor retrieval (HnE $\rightarrow$ MP and MP $\rightarrow$ HnE) with 20\% and 80\% correspondences. Fig. 3(a) depicts the qualitative results of 3 cross-sections and corresponding retrived modality-couterpart images. The normal (N: left column), late fibroatheroma (FA: middle column), and pathological intimal thickening (PIT: right column) plaques  have been successfuly retrived on the top 3 ranking results and only two are incorrectly fetched (red boxed) as the fourth neighbors. Such a retrieval tool will significantly improve histopathologist's ability to make reliable and fast decision.

\begin{figure}[t]
\centering
\includegraphics[width = \textwidth]{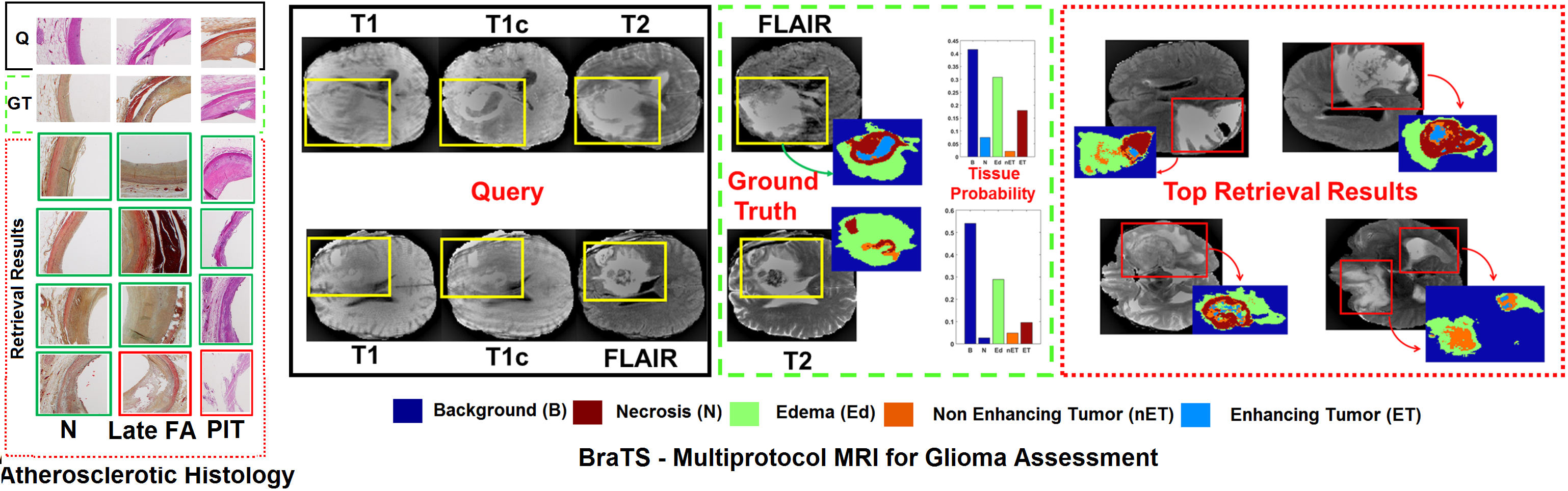}
\caption{Qualitative results for (a) atherosclerotic histology and (b) BraTS datasets. Query (Q) image along with `ground truth' (GT) and top fetched cross-modal images (\textcolor{green}{green} box - similar annotation as Query and \textcolor{red}{red} box - dissimilar annotation ).}
\label{fig:illustrativeEx}
\end{figure}

\noindent
\textbf{Regression for glioma assessment through cross-modal retrieval}:
We pose the retrieval task defined over a publicly available multi-protocol MR dataset for Glioma assessment (BraTS)~\cite{menze2015}. In total, 2170 variable sized ROIs($> 6$ cm $\times$ 6 cm) encompassing tumors were selected across four modalities\footnote{The modalities are pre-aligned to the standard Brainweb space and re-sampled with an isotropic resolution of 1 mm.}. The associated ground-truth segmentations are used to construct a five-element tissue probability vector (say, $c_{p}$) descriptive of the tumor's composition (in Fig. 3(b)).  

\noindent
\textit{Results}: We consider testing scenarios where a triad of modalities are available as a query (say $\mathbf{x}_{q}$) and the goal is to retrieve the ROIs from the complementary modality that are similar in tissue composition. In this case, the retrieval performance is measured using the average log root mean square deviation (alogRMSD) between the `ground-truth' tissue probability of the query ($c_{p}(\mathbf{x}_{q})$) to that of the top $k$ nearest cross-modal neighbors over $n_{t}$ test samples: $\text{alogRMSD} = -\log\left ( \left ( 1/n_{t} \right )\sum_{q = 1}^{n_{t}}\sqrt{\sum_{j = 1}^{k}\left ( c_{p}(\mathbf{x}_{q}) - c_{p}(\mathbf{x}_{q}^{k}) \right )^{2}/k} \right )$. We create four independent test scenarios varying the query triads as: \textbf{S1}: T1, T1c, T2 $\rightarrow$ FLAIR; \textbf{S2}: T1, T2, FLAIR $\rightarrow$ T1c; \textbf{S3}: T2, T1c, FLAIR $\rightarrow$ T1 and \textbf{S4}: T1, T1c, FLAIR $\rightarrow$ T2. Similar to previous experiment, we used AlexNet~\cite{krizhevsky2012} for generating the query modality features (and na\"{i}vely concatenate them), VGG-F~\cite{chatfield2014} for the target modality, and finally removed feature correlations through PCA. Fig.~\ref{fig:resultsKCurve} shows the alogRMSD \textit{vs.} scope ($k$) curves for two comparative settings of 20\% and 80\% correspondences, evaluating the overall performance. Fig.~\ref{fig:illustrativeEx}(b) also demonstrates the qualitative retrieval results for two ROIs under \textbf{S1} and \textbf{S4}, where retrieved cross-modal neighbors (FLAIR(/T2) modality using a  T1, T1c, T2(/FLAIR) query triad) exhibit significant tissue compositional similarity with that of the query image. Notably, the expression of necrotic core (with enhancing tumor) engulfed by edema is consistent across retrieved neighbors. 

\noindent
\textbf{Observations}:
From Fig.~\ref{fig:resultsKCurve}, we observe that the two proposed variants of C$\text{M}^{2}$L present a trend of consistently higher performance against comparative methods, substantiating the superiority of preseving joint global and local geometries. The performances of majority of methods are improved as degree of correspondences is increased from 20\% to 80\%. In case of C$\text{M}^{2}$L, this can be attributed to the better approximation of cross-modal affinity through given corresponding `links' and hence making cross-modal data comparable in the latent space. Meanwhile, in majority of the cases, C$\text{M}^{2}$L-I shows improved performance over C$\text{M}^{2}$L-F due to the inherent non-linear flexibility of mapping data onto the embedding space.  
The LCFS performance is closest to C$\text{M}^{2}$L as it discovers discriminative common latent space features, which make the embedding compact and effective. Additionally, despite considering global geometries while generating embedding, the MA-I underperformed, because, the Euclidean distance dissimilarity metric is not suitable for representing semantic similarity between instances.

\section{Conclusions}

We proposed C$\text{M}^{2}$L for effective cross-modal retrieval in which heterogeneous gap between cross-modal feature spaces is bridged by embedding instances into a metric-comparable latent space. In C$\text{M}^{2}$L, both local and global geometries are respected simultaneously using limited number of corresponding instances. To the best of our knowledge, this is the first cross-modal medical image retrieval technique and we will extend it to multimedia (text+image) datasets in future.

\end{document}